\RequirePackage[table]{xcolor}
\documentclass[journal,twoside,web]{ieeecolor_official}
\usepackage{tmi}
\usepackage{amsmath,amssymb,amsfonts}
\usepackage{algorithmic}
\usepackage{graphicx}
\usepackage{algorithm,algorithmic}
\usepackage{textcomp}
\usepackage{cite}
\usepackage{romannum}
\usepackage{multirow}
\usepackage{makecell}
\usepackage{textcomp}
\usepackage{fontawesome}
\usepackage{hyperref}
\usepackage{bbding}

\makeatletter

\newcommand{\Rmnum}[1]{\expandafter\@slowromancap\romannumeral #1@}
\makeatother

\definecolor{lightblue}{RGB}{214, 228, 255}

\def\BibTeX{{\rm B\kern-.05em{\sc i\kern-.025em b}\kern-.08em
    T\kern-.1667em\lower.7ex\hbox{E}\kern-.125emX}}
\begin{document}

\title{Pos4CHD: A Positive-Free Mask-Guided Counterfactual Synthesis Framework for Fetal Congenital Heart Defect Analysis}
\author{
    Wenjie Xuan,
    Yapeng Li,
    Hang Su,
    Chen Cheng,
    Juhua Liu, \IEEEmembership{Member, IEEE}, \\
    Sheng Zhao, 
    Ziming Huang,
    and Bo Du, \IEEEmembership{Senior Member, IEEE}
    \thanks{This work was supported in part by the National Key Research and Development Program of China under Grant 2023YFC2705701, in part by the National Natural Science Foundation of China under Grants 62225113, 625B2132 and U23B2048, in part by the Science and Technology Major Project of Hubei Province under Grants 2025BCB026, and in part by the China Birth Defect Intervention and Rescue Foundation and Beijing ByteDance Charity Foundation under 2026CFBD-CKZN-005. \textit{Corresponding author: Juhua Liu (e-mail: liujuhua@whu.edu.cn)}}
    \thanks{Wenjie Xuan, Yapeng Li, Hang Su, Juhua Liu and Bo Du are with the School of Computer Science, Wuhan University, Wuhan, China. (\Envelope: \{dreamxwj, yapengli, hangsu, liujuhua, dubo\}@whu.edu.cn).}
    \thanks{Chen Cheng, Sheng Zhao and Ziming Huang are with Maternal and Child Health Hospital of Hubei Province, Wuhan, China. (\Envelope: \{chengchen 777, zhao1015568495\}@163.com, huangziming@hbfy.com).}
}
\maketitle

\begin{abstract}
    The scarcity of positive cases remains a major bottleneck in developing automated diagnostic systems for fetal Congenital Heart Defects~(CHDs), particularly for rare subtypes. 
    Although normal cases are accessible with effort, positive cases are hard to collect due to low incidence rates and privacy concerns, causing severe class imbalance or absence of positive training samples. 
    Existing works resort to synthesizing positives with conditional generative models, but they are task-specific and require positive training samples, hindering application for fetal CHDs. 
    To this end, we propose a positive-free mask-guided counterfactual generation framework for fetal echocardiography, namely Pos4CHD, which enables realistic, high-quality synthesis of pseudo-positive samples from semantic masks, with only normal samples at hand. Technically, we first develop FE-Gen, a mask-guided generative model tailored for fetal heart ultrasound synthesis that mitigates histogram drift and enhances image-mask spatial alignment. Building on FE-Gen, Pos4CHD employs a mask-to-image generation paradigm for CHD synthesis. It first manipulates semantic masks to simulate fetal-heart structural anomalies and then renders anatomically consistent images.
    Extensive experiments on in-house datasets, together with expert evaluations, demonstrate the visual fidelity~($-12.97$ FID) and anatomical consistency~($+3.96\%$ DSC) of synthesized images. Further validations across diverse real-world CHD diagnosis tasks exhibit promising results and consistent increments, including zero-shot classification~(achieving $93.25\%$ ACC on EbA and $72.92\%$ ACC on TAPVC), lesion detection~($+6.95\%$ mAP50 on VSD), and anatomy segmentation~($+10.49\%$ DSC on fetal heart) compared with baselines.
    Codes and models are publicly available on \href{https://github.com/DREAMXFAR/Pos4CHD}{GitHub}.
\end{abstract}


\section{Introduction}
\label{sec:introduction}

    \begin{figure}[t]
        \centering
        \includegraphics[width=0.95\linewidth]{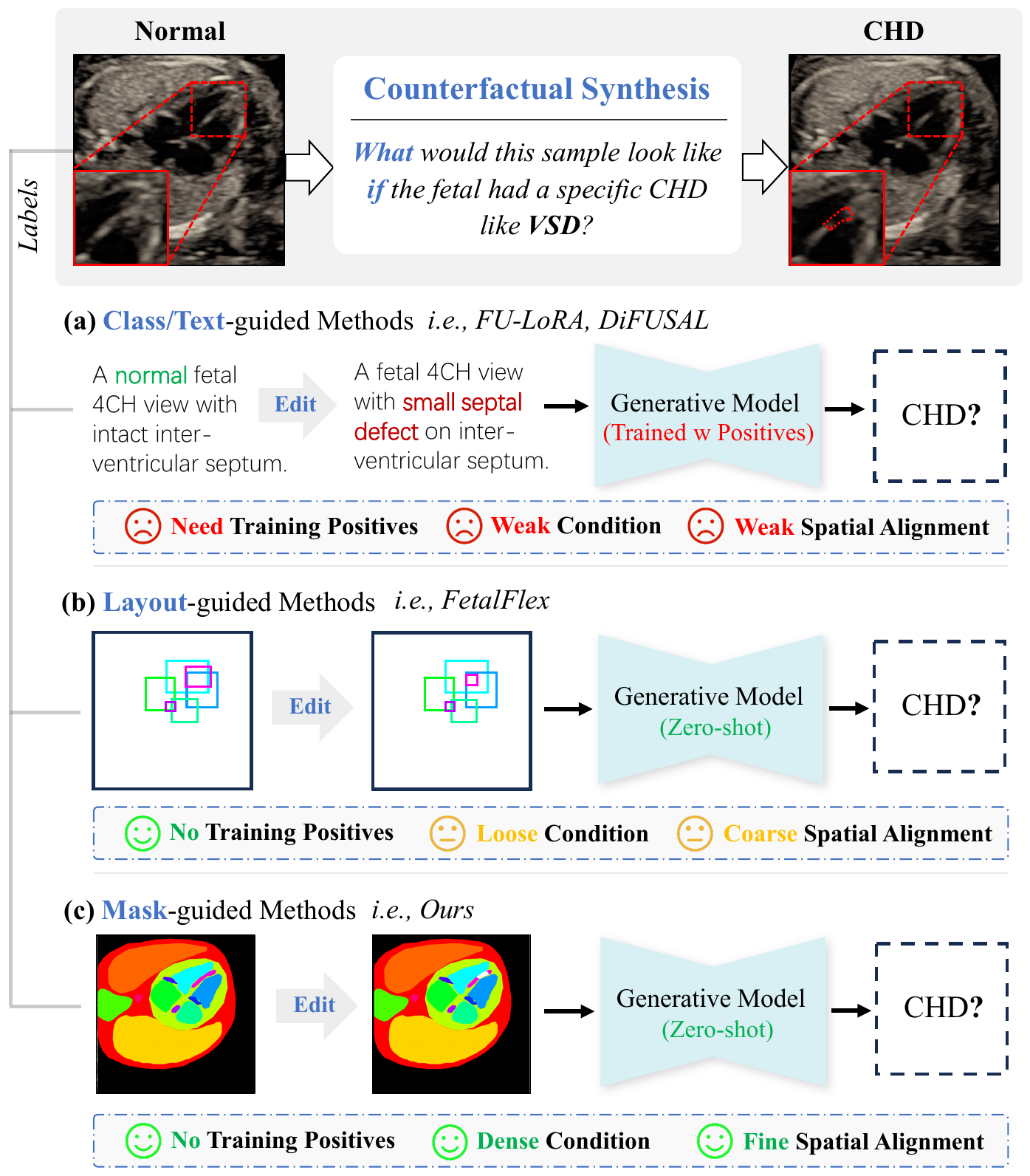}
        \caption{Paradigm comparison for (a) class/text-, (b) layout-, (c) mask-guided medical counterfactual synthesis. Since CHDs are structural anomalies, we employ a mask-guided paradigm for fetal CHD synthesis without positive training samples, providing dense structural guidance, fine spatial alignment, and interpretability for downstream analysis.}
        \vspace{-1.2em}
        \label{fig:figure-1}
    \end{figure}

    \IEEEPARstart{F}{etal} congenital heart defects~(CHDs), characterized by heterogeneous structural anomalies, are one of the leading causes of infant mortality, which affects approximately $0.8$$\sim$$1.0\%$ of live births~\cite{carvalho2023isuog, liu2019global}. Fetal echocardiography, {\it i.e.}, fetal heart ultrasound, is the primary tool for prenatal examination due to its radiation-free, low-cost, and real-time nature. Clinicians typically examine fetal heart anatomical structures using standardized cardiac views~\cite{moon2023guidelines} like the Four-Chamber~(4CH) view, Three-Vessel and Trachea~(3VT) view, {\it etc.}, which provide essential and vital references for cardiac function analysis. This practice helps early detection and diagnosis of various CHD subtypes, including Ebstein's Anomaly~(EbA), Total Anomalous Pulmonary Venous Connection~(TAPVC), Ventricular Septal Defect~(VSD), {\it etc.}, contributing to improved worldwide pregnancy healthcare~\cite{sun2021prenatal}.
    
    However, compared with existing AI systems for MRI, CT, and X-rays~\cite{ 11659599}, AI-assisted fetal CHD detection and analysis remains underdeveloped due to severe data scarcity, which stems from: (1) the acquisition of standardized fetal cardiac views requires expertise and experience, (2) varied view quality across operators and examinations, and (3) cardiac anatomical annotation is labor-intensive and requires expertise, resulting in few high-quality public datasets. More importantly, given the wide range of CHD subtypes ($\geq$$27$), although normal cases are accessible through clinical efforts, positive cases for rare CHDs are often out of reach owing to low incidence rates ($<$$1/10,000$)~\cite{liu2019global} and privacy constraints. 
    This positive-sample scarcity hinders the training of downstream models, leading to severe class imbalance and potential generalization issues~\cite{fiorentino2023review}. 

    Recent advances in generative models, {\it e.g.}, diffusion models, offer a promising solution to medical data scarcity via image synthesis. Existing studies focused on improving image fidelity~\cite{Dom_Diffusion_MICCAI2024,guo2025maisi} and anatomical plausibility~\cite{xu2024medsyn,mao2025medsegfactory} across different imaging modalities, and adapting generative foundation models for medical applications~\cite{anatodiff, wu2025mrgen,zhang2025generative}. 
    Conditional generative models, {\it e.g.}, Stable Diffusion~(SD)~\cite{rombach2022stablediffusion} and ControlNet~\cite{zhang2023controlnet}, further enable task-specific conditional guidance to synthesize images with desired properties. Practices (see Fig.~\ref{fig:figure-1}) include class labels or text prompts~\cite{wang2025toward, perez2024radedit} for anatomical description, layouts~\cite{fetalflex} for lesion position and size, segmentation masks~\cite{zhao2026maisi,konz2024anatomically} for structure and shape, with optional reference images~\cite{min2025instructx2x, 10.1007/978-3-031-73281-2_16} to provide extra context for local variations. Such methods can produce paired image-condition data for augmentation purposes and benefit various downstream tasks including classification, segmentation, and multi-center generalization~\cite{mao2025medsegfactory, oliveras2026anatomically}. Recent counterfactual synthesis explores generating positive cases, such as tumors~\cite{chen2024towards}, thyroid cancers~\cite{dai2025improving}, breast lesions~\cite{chen2025subtyping}, {\it etc}. 

    Despite these advances, directly applying such approaches to fetal CHD synthesis remains challenging. 
    First, most works are designed for specific tasks and limited scenarios. We need to redesign a framework for fetal CHD synthesis, where the complicated cardiac anatomy and heterogeneous CHD subtypes raise the complexity. Second, current methods~\cite{chen2024towards, dai2025improving, chen2025subtyping} require positive samples to learn lesion-specific textures and structures, implicitly assuming the availability of positive cases. But this assumption is violated because of the absence of fetal CHD samples, particularly for rare subtypes.     
    Third, existing works lack fine-grained control over synthesized images. As shown in Fig.~\ref{fig:figure-1}, given that CHDs are mostly structural anomalies, the class labels and text prompts offer limited spatial control, where the anatomical plausibility of synthesized positives depends on generative priors and is hardly guaranteed. Though bounding boxes offer locations and enable simulation of position and size variations, they still fall short in depicting morphology and fine-grained variations, thus insufficient to distinguish anatomical differences across CHD subtypes.
    Therefore, we raise the question: {\it can we design a unified framework to synthesize realistic and anatomy-consistent samples of diverse CHD subtypes for downstream analysis, with only negative samples at hand?} 
    
    To this end, we propose {\bf Pos4CHD}, a positive-free mask-guided counterfactual synthesis framework for fetal CHD analysis. Firstly, considering the absence of publicly available datasets, we collect $\sim$8k fetal 4CH views and $\sim$$500$ expert-annotations covering 14 key anatomical structures (Sec.~\ref{sec:data_collection}), and train {\bf FE-Gen} (Sec.~\ref{sec:FE-Gen arch}), a mask-conditioned generator for fetal echocardiography, which features: 1) learnable semantic queries that enhance the spatial alignment between synthesized images and conditioning masks, and 2) a regularization term that mitigates the histogram drift when adapting diffusion models for ultrasound with improved image fidelity.
    Building on FE-Gen, Pos4CHD (Sec.~\ref{sec:FreePos}) utilizes a mask-to-image generation strategy to synthesize fetal CHDs from normal samples. Specifically, it first manipulates cardiac masks to simulate structural anomalies with respect to clinical prior knowledge, realizing rule-based counterfactual mask generation for fetal CHDs. Then, FE-Gen renders corresponding echocardiographic images guided by masks, yielding spatially aligned pseudo-positive samples. These synthesized samples are subsequently utilized to augment downstream training data after quality checking. 
    Experiments and expert evaluations demonstrate the image fidelity and anatomical consistency of FE-Gen (Sec.~\ref{sec:exp_generation}).
    Extensive experiments on real-world scenarios, including zero-shot EbA/TAPVC classification, VSD detection, and fetal cardiac anatomy segmentation, verify the effectiveness of Pos4CHD, which not only enables training models without real positives but also consistently outperforms the baseline (Sec.~\ref{sec:exp_downstream}). Notably, models trained solely on synthesized data can generalize well to real samples. 
    
    Our contributions are summarized as follows:
    \begin{enumerate}
        \item[1)] We develop {\bf Pos4CHD}, a positive-free mask-to-image counterfactual generation framework to tackle the scarcity of fetal CHD cases, especially rare subtypes, which offers pseudo-positive image-mask pairs for diverse CHDs through rule-based counterfactual mask manipulation and mask-guided fetal ultrasound synthesis.

        \item[2)] We collect normal fetal 4CH-view datasets with rich pixel-level anatomical annotation. Based on the collected data, we train {\bf FE-Gen}, a mask-guided generative model for fetal echocardiography, realizing improved ultrasound fidelity and advanced image-mask alignment. 
        
        \item[3)] Experiments together with expert evaluations demonstrate the effectiveness of our synthesized samples for data augmentation across various downstream tasks, bringing consistent improvements on zero-shot CHD classification, lesion detection, and fetal heart anatomy segmentation. Detailed ablations and discussions provide valuable insights to the community on how to obtain and utilize the synthesized positive samples with our method. 
    \end{enumerate}

\section{Related Work}
\label{sec:related_work}
    \subsection{Medical Image Generation}
        Medical image generation has been widely explored to alleviate the scarcity, limited accessibility, and high acquisition costs of data across different modalities~\cite{11659599}. Following early GAN-based approaches, diffusion models have become the primary choice for high image fidelity and training stability nowadays. Recent studies moved beyond visual realism toward medical-specific challenges, including anatomical fidelity~\cite{xu2024medsyn, konz2024anatomically,mao2025medsegfactory}, few-shot adaptation~\cite{wu2025mrgen,zhang2025generative}, and privacy preservation.
        Recently, controllable generation has gained increasing attention for downstream clinical applications. By incorporating conditions like patient metadata~\cite{wang2025toward}, modalities and styles~\cite{zhan2024medm2gunifyingmedicalmultimodal}, pathological states~\cite{abir2026hybrid}, and anatomical layouts or masks~\cite{zhao2026maisi, chen2024towards}, conditional generative models enable targeted synthesis of samples with desired characteristics. Such synthetic samples are privacy-friendly, valuable for mitigating class imbalance, enriching data diversity, and improving the robustness and generalization of medical AI models. 
    
    \subsection{Ultrasound Image Generation}
        Ultrasound image generation attracts growing attention due to its broad clinical use and distinctive imaging characteristics. Recent studies incorporated physical priors~\cite{Dom_Diffusion_MICCAI2024} and structural constraints~\cite{9682549, wen2026synthesizing} to improve fidelity. Controllable ultrasound generation has also been explored for downstream applications, such as augmenting image-mask pairs for segmentation~\cite{stojanovski2023echo} and enriching data diversity~\cite{van2025generative, yazdani2025flow}. 
        
        Fetal ultrasound generation is a crucial research direction. 
        FU-LoRA~\cite{wang2024generative} and DiFUSAL~\cite{arjemandi2025difusal} fine-tuned SD for text-conditioned fetal ultrasound synthesis to enhance plane classification and gestational age prediction. 
        FVEDM~\cite{QIAO2026128939} combines image inpainting with semantic masks and edge guidance to introduce localized variations to 4CH views, thereby enriching appearance diversity for CHD classification. 
        FetalFlex~\cite{fetalflex} employs masked inpainting with anatomical layouts and global prompts for multi-plane fetal ultrasound synthesis. 
        Although similar to FetalFlex in motivation, we adopt a fine-grained mask-to-image paradigm for ultrasound synthesis to achieve precise spatial control, and we mitigate the histogram shift to improve fidelity. Besides, Pos4CHD targets the more challenging fetal CHD synthesis for downstream analysis, where fine-grained anatomical fidelity is critical for distinguishing heterogeneous subtypes with complex cardiac structures.
    
    \subsection{Counterfactual Synthesis}
        Counterfactual synthesis produces hypothetical images by altering specific attributes of an existing sample while preserving its remaining characteristics, corresponding to \textit{what-if} interventions as illustrated in Fig.~\ref{fig:figure-1}. Such counterfactual samples are valuable for model robustness evaluation and interpretation, anomaly detection, and data augmentation. This paper focuses on augmenting positive cases. Existing works typically rely on conditional generation or image editing~\cite{min2025instructx2x, 10.1007/978-3-031-73281-2_16} to synthesize positives, guided by class labels or text prompts~\cite{yeganeh2025latent, liu2024controllable}, or spatial conditions like edges, layouts, and masks~\cite{Li2026integrating, wu2024improving} in Fig.~\ref{fig:figure-1}. However, disease-specific synthesis usually requires pathological samples or supervision to learn appearance and structural characteristics associated with the target disease. For example, Tiger~\cite{dai2025improving} performs text-guided synthesis to augment rare thyroid cancer subtypes but requires cancer samples to train the generative model. This dependency restricts applicability to diseases like fetal CHDs, where positive cases are scarce and infeasible to collect.
        Therefore, Pos4CHD targets subtype-specific fetal CHD synthesis, characterized by structural anomalies, using only normal samples. Combining rule-based mask manipulation and mask-guided ultrasound synthesis, Pos4CHD enables fine-grained modeling and synthesis of heterogeneous CHD subtypes with verifiable semantic masks for clinical applications.

\section{Methodology}
\label{sec:method}

    \begin{figure*}[t]
        \centering
        \includegraphics[width=0.97\linewidth]{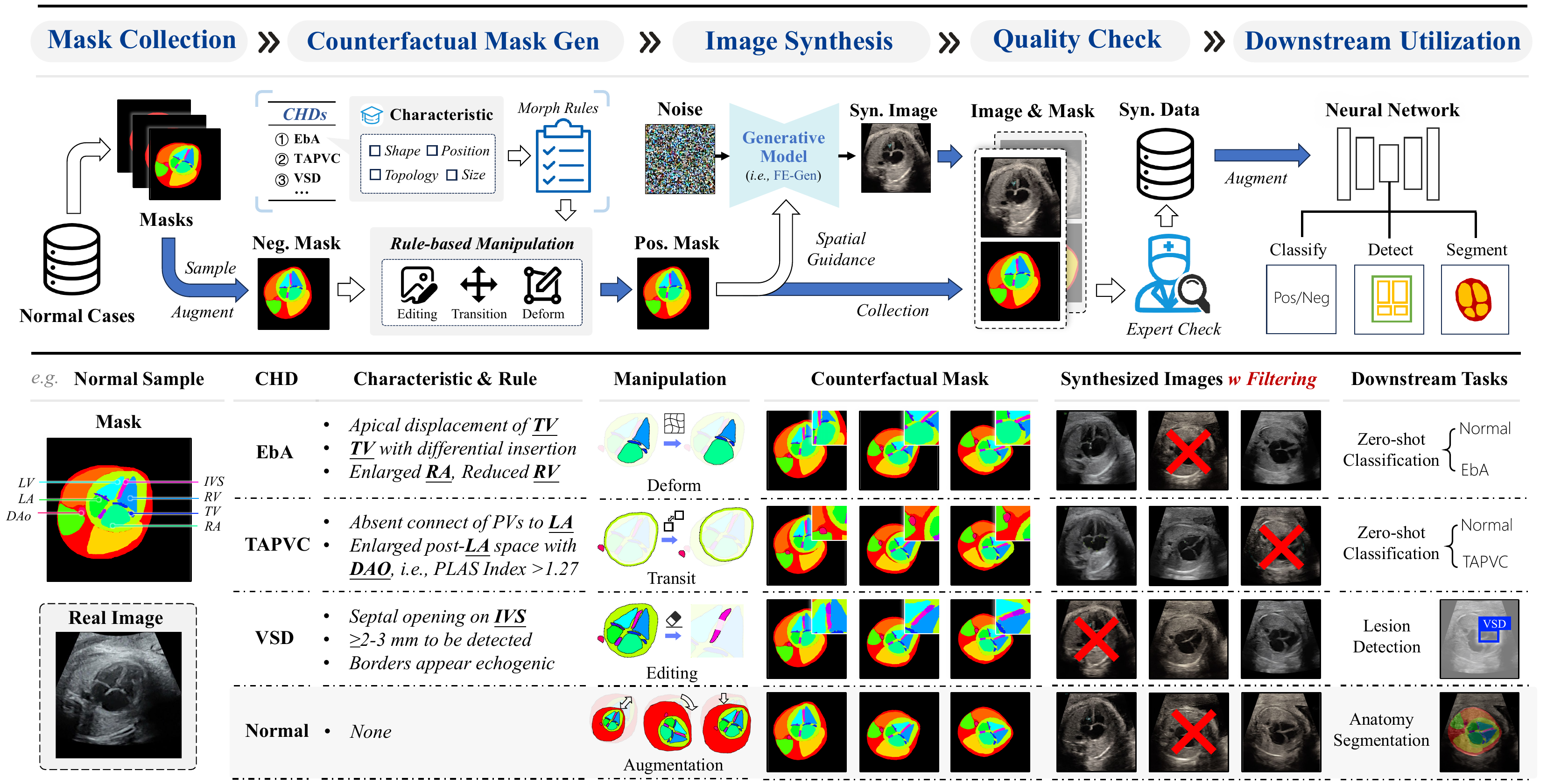}
        \caption{Overview of our Pos4CHD framework, which synthesizes counterfactual positive samples for fetal CHD analysis. It comprises five steps: (1) mask collection from normal samples, (2) rule-based counterfactual mask generation, (3) pseudo-positive image synthesis with our proposed FE-Gen, (4) quality check and data filtering, and (5) downstream utilization including classification, detection, and segmentation.}
        \vspace{-1.2em}
        \label{fig:FreePos}
    \end{figure*}

    \subsection{Preliminaries}
    \label{sec:preliminary}
        \subsubsection{Diffusion Models}
        The diffusion probabilistic model has been the primary choice for image generation nowadays. 
        It defines a Markov chain, where images are first corrupted by a series of Gaussian noise~({\it forward process}), and then it samples synthesized images from a learned distribution by starting from random Gaussian noise ({\it backward process}). 
        Supposing a sample $\boldsymbol{x}_0$$\sim$$ q(\boldsymbol{x})$, the forward diffusion process adds increasing gaussian noise $\epsilon$ to the clean data $\boldsymbol{x}_0$ according to timestep $t$$\in $$T$, which is defined as $q(\boldsymbol{x}_t|\boldsymbol{x}_{t-1})$$=$$\mathcal{N}(\boldsymbol{x}_t; \alpha_t \boldsymbol{x}_{t-1}, \beta_t \boldsymbol{I})$, where $\alpha_t, \beta_t\in [0,1]$ denote the noise schedule for image corruption and satisfy $\alpha_t ^2+\beta_t^2$$=$$1$ for variance preservation. This process can be written and re-parameterized as one-step noising, $\boldsymbol{x}_t = \sqrt{\overline{\alpha}_t}\boldsymbol{x_0} + \sqrt{1-\overline{\alpha}_t}\epsilon$, $\epsilon$$\sim$$\mathcal{N}(\boldsymbol{0}, \boldsymbol{I})$, 
        where $\overline{\alpha}_t$$=$$\prod_1^t \alpha_t$. 
        The backward sampling process starts from a random Gaussian noise $\boldsymbol{x}_T$, which is reversed back towards the original data distribution $q(\boldsymbol{x})$ via iterative denoising, formatted as $p_\theta(\boldsymbol{x}_{0:T})$$=$$p(\boldsymbol{x_T})\prod_{t=1}^T p_\theta(\boldsymbol{x}_{t-1}|\boldsymbol{x}_t)$, where $p_\theta(\boldsymbol{x}_{t-1}|\boldsymbol{x}_t)$$=$$\mathcal{N}(\boldsymbol{x}_{t-1};\mu_\theta(\boldsymbol{x_t}, t), \beta_t\boldsymbol{I})$. Since the exact calculation of $\hat{\mu}_\theta$ is intractable, it is approximated by a network $\hat{\epsilon}_\theta(\boldsymbol{x_t}, t)$, optimized by ELBO, and formatted as, 
        \begin{equation}
            \hat{\mu}_\theta(\boldsymbol{x_t}, t)=\frac{1}{\sqrt{\alpha}_t}(\boldsymbol{x}_t-\frac{\beta_t}{\sqrt{1-\overline{\alpha}_t}} \hat{\epsilon}_\theta(\boldsymbol{x_t}, t)).
        \end{equation}
        Therefore, the final training objective can be further simplified as a denoising target, namely $\epsilon$-prediction, formatted as, 
        \begin{equation}
            \mathcal{L}_{denoise} = \mathbb{E}_{\boldsymbol{x}_0,t,\epsilon}|| \epsilon-\hat{\epsilon}_{\theta}(\boldsymbol{x}_t, t) ||_2^2.
        \end{equation} 

        To reduce the computational cost, SD~\cite{rombach2022stablediffusion} moves the diffusion process to a latent space constructed by VAE, where the image latent $\boldsymbol{x}$$=$$\mathcal{E}(I)$ and the reconstructed image $\hat{I}$$=$$\mathcal{D}(\boldsymbol{x})$, where an encoder $\mathcal{E}(\cdot)$ and decoder $\mathcal{D}(\cdot)$ are shown in Fig.~\ref{fig:FE-Gen}, namely Latent Diffusion Models. This paradigm has achieved impressive results in natural image generation tasks, and researchers have explored its applications in medical image generation for the long-standing issue of data scarcity. 

        \subsubsection{Spatially Controllable Image Generation}  
        The representative ControlNet~\cite{zhang2023controlnet} introduces adapters to SD to enable task-specific spatial control over image synthesis, {\it e.g.}, edges, semantic masks, and depths, which shows impressive results on customized image generation and application. Supposing one encoder block $F(\cdot; \theta)$ of SD, it accepts the input feature $\boldsymbol{x}$ and output $\boldsymbol{y}$$=$$F(\boldsymbol{x};\theta)$. ControlNet makes a trainable copy $F(\cdot; \theta^\prime)$ to inject condition $\boldsymbol{c}$ into the model with zero-initialized convolutions $Z(\cdot;w)$ for stability, which is formulated as, 
        \begin{equation}
            \boldsymbol{y_c} = F(\boldsymbol{x};\theta) + \lambda * Z(F(\boldsymbol{x}+Z(\boldsymbol{c};w_1); \theta^\prime); w_2), 
            \label{eq:add-condition}
        \end{equation}
        where $\boldsymbol{y_c}$ is the output feature modulated by spatial conditions. The conditioning scale $\lambda$ is introduced to adjust control strength during inference. The training objective for conditional generative models is formatted as, 
        \begin{equation}
            \mathcal{L}_{denoise} = \mathbb{E}_{\boldsymbol{x}_0,t,\boldsymbol{c},\epsilon}|| \epsilon-\hat{\epsilon}_{\theta}(\boldsymbol{x}_t, t, \boldsymbol{c}) ||_2^2.
        \end{equation}

    \subsection{Pos4CHD: A Positive-free Mask-to-image Counter-  factual Generation Framework for Fetal CHD Analysis}
    \label{sec:FreePos}
        Given the scarcity or absence of fetal CHD cases, directly learning disease-specific appearance and structures from positives is infeasible. Thus, we formulate CHD synthesis as mask-to-image counterfactual generation, where semantic masks explicitly encode CHD-related anatomical changes and guide the synthesis of positive samples. This scheme supports generation of diverse CHD subtypes within a unified framework. 
        
        Our design is motivated by three observations: 
        (1) Most CHDs are presented as cardiac structural anomalies. Semantic masks therefore provide an explicit and interpretable representation of cardiac anatomy and morphology, as a unified proxy for CHD analysis; 
        (2) Semantic masks provide dense, fine-grained spatial control and can be manipulated easily, flexibly, and reliably, which enables CHD simulation with anatomical consistency according to clinical priors; 
        (3) Counterfactual masks offer explicit references for verifying the anatomical correctness of synthesized samples while supplying pixel-level supervision for downstream tasks at no extra annotation cost.
        
        The pipeline of Pos4CHD is shown in Fig.~\ref{fig:FreePos}, which consists of five steps: (1) normal mask collection, (2) counterfactual mask manipulation, (3) pseudo-positive image synthesis, (4) quality check and data filtering, and (5) downstream utilization. The implementation of each step is depicted as follows.
        
        We begin with a set of normal fetal cardiac masks, denoted as $\mathcal{M}$$=$$\{\mathbf{m}_i\}_{i=1}^N$, obtained through manual annotation, pretrained segmentation models, or atlas animations. These masks serve as the basis for counterfactual sample synthesis. 
        Given a target fetal CHD subtype $c$$\in$$\mathcal{C}$, we first design morphological transformation rules $R(\cdot, c; \psi_c)$ based on clinical prior knowledge, thereby simulating plausible counterfactual masks with subtype-specific characteristics through operations like translations, editing, free-form deformation, {\it etc.}, where $\psi_c$ denotes category-specific transformation hyperparameters. 
        Specifically, this paper focuses on three representative heterogeneous subtypes, {\it i.e.,} EbA, TAPVC, and VSD. Their key morphological properties~\cite{isuog_abnormal_4ch} and corresponding manipulation rules are summarized in Fig.~\ref{fig:FreePos}.
        Next, we apply customized transformation $R(\cdot, c; \psi_c)$ to the normal mask $\mathbf{m}$, yielding counterfactual cardiac mask $\overline{\mathbf{m}}$. This clinical-knowledge-driven counterfactual mask manipulation is formulated as, 
        \begin{equation}
            \overline{\mathbf{m}} = R(\mathbf{m}, c; \psi_c).
            \label{eq:mask_transform}
        \end{equation}
        
        Given the counterfactual mask $\overline{\mathbf{m}}$, we employ a mask-guided conditional generative model $\mathcal{G}(\epsilon, \mathbf{m})$ to generate spatially aligned, realistic fetal echocardiographic images, where the synthesis of the pseudo-positive image $\overline{I}$ is formulated as, 
        \begin{equation}
            \overline{I} = \mathcal{G}(\epsilon, \overline{\mathbf{m}}),\ \epsilon\sim \mathcal{N}(\boldsymbol{0},\boldsymbol{I}), 
            \label{eq:image gen}
        \end{equation}
        where we employ a customized generator, FE-Gen (see Sec. \ref{sec:FE-Gen}), for advanced image fidelity and image-mask alignment for fetal ultrasound synthesis. Following the pipeline in Fig.~\ref{fig:FreePos}, we can produce diverse candidate pseudo-positive image-mask pairs $(\overline{I},\overline{\mathbf m})$ for CHD subtypes, owing to the randomness of mask manipulation, augmentations, and image generation.
        
        Then, we inspect and filter low-quality synthesized samples $(\overline{I},\overline{\mathbf m})$ manually or with pretrained segmentation models to construct a high-quality synthetic dataset $\mathcal{\overline D}$$=$$\{(\overline{I}_i,\overline{\mathbf m}_i)\}_{i=1}^{N^\prime}$ of pseudo-positives with pixel-level annotations. This is because the stochastic nature of diffusion sampling may introduce artifacts~\cite{wang2025generative} or unfaithful CHD samples, particularly in medical scenarios requiring precise anatomical fidelity, as validated in Sec.~\ref{sec:valid_filtering}. The resulting image-mask pairs can be readily converted into class labels or bounding boxes, supporting diverse real-world CHD analysis tasks.

        \begin{figure}[b]
            \centering
            \vspace{-0.4em}
            \includegraphics[width=0.94\linewidth]{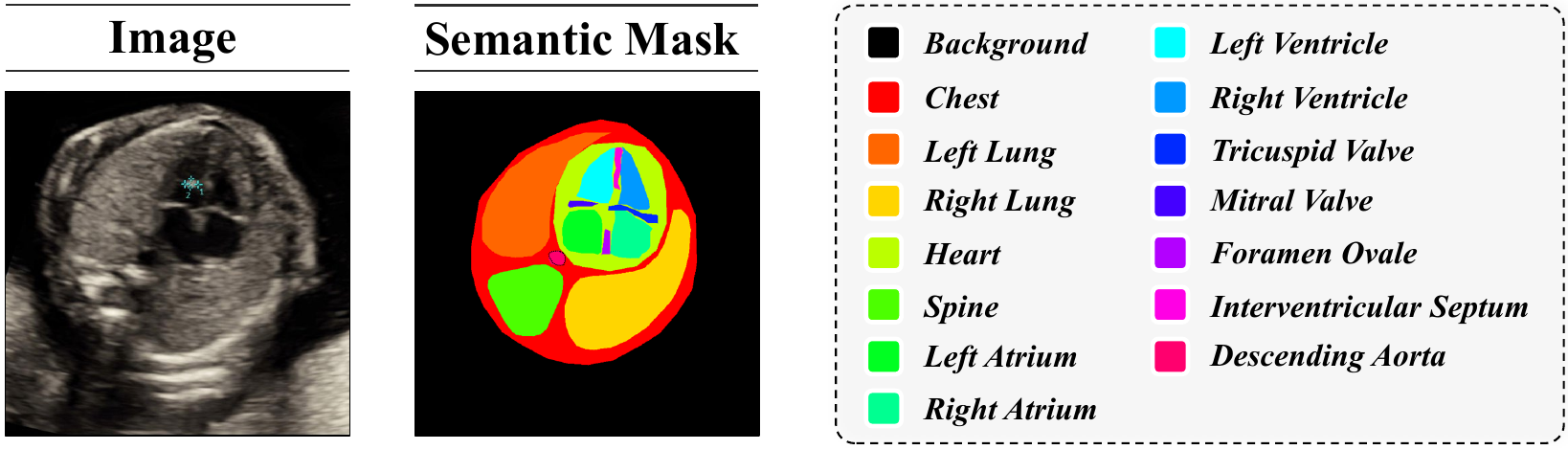}
            \caption{Example fetal cardiac 4CH view with 14 anatomical annotations.}
            \label{fig:heart annotation}
        \end{figure}

    \subsection{FE-Gen: A Mask-guided Controllable Generation Model for Fetal Echocardiography}
    \label{sec:FE-Gen}    
        As discussed above, Pos4CHD enables CHD synthesis using only negative cases, but its performance relies on the quality of mask-to-image generator $\mathcal{G}(\cdot)$, where two key factors matter: (1) {\it image fidelity}, which decides how closely the synthesized fetal echocardiographic images match the real ones in visual appearance and data distribution; and (2) {\it spatial alignment} between each synthesized image and conditioning mask, which determines the faithfulness of CHD-specific characteristics and mask accuracy. Poor spatial alignment would introduce noisy labels, thereby degrading downstream performance and data efficiency. Therefore, an effective generator should produce realistic fetal heart ultrasound with precise image-mask alignment.
        The implementation details are as follows.
    
        \subsubsection{Data Collection} 
        \label{sec:data_collection}
            Due to the lack of public fetal echocardiography data, we collect an in-house dataset of fetal heart ultrasound from the hospital, namely \texttt{WHU-4CH-8k}. This dataset contains $7,974$ fetal heart 4CH views from $2,855$ patients acquired between 2017 and 2023, mostly ranging from 21-23 and 28-30 gestational weeks, after plane quality checks~\cite{ultrasound2022chinese}. This unlabeled dataset is utilized to train the fetal ultrasound generator. 
            We further select $517$ high-quality samples and annotate 14 key anatomical structures with clinical experts, namely \texttt{WHU-4CH-Seg517}.
            As shown in Fig.~\ref{fig:heart annotation}, the rich semantic annotations on 4CH views provide anatomical and spatial guidance to the generator, supporting simulation of CHD subtypes associated with these structures. More details, including data splits and resolution, are provided in Sec.~\ref{sec:settings}.

        \begin{figure}[t]
            \centering
            \includegraphics[width=0.96\linewidth]{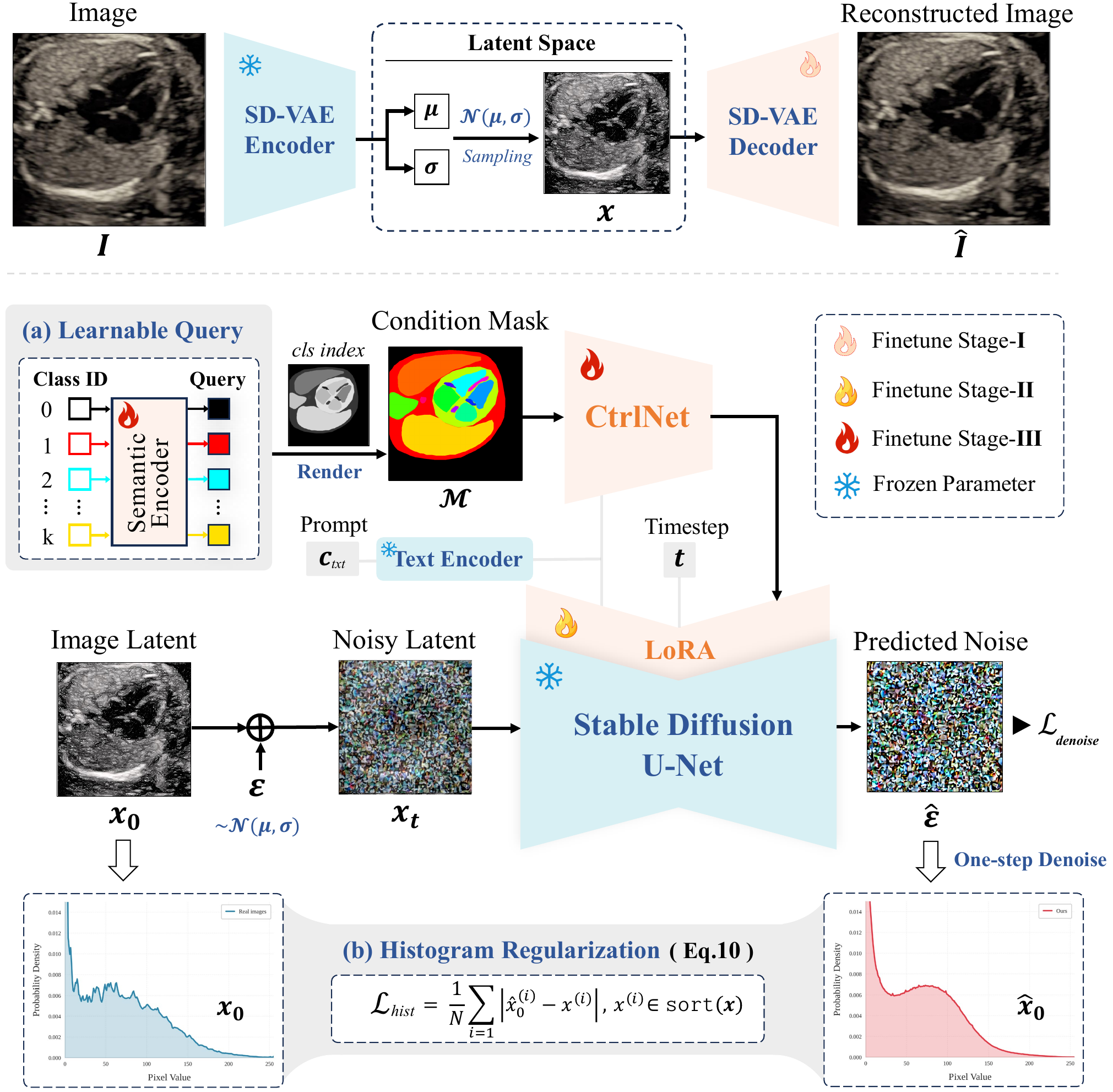}
            \caption{Architecture of FE-Gen: (a) Learnable semantic queries for spatial alignment, and (b) Histogram regularization for ultrasound fidelity.}
            \vspace{-1.6em}
            \label{fig:FE-Gen}
        \end{figure}

        \subsubsection{FE-Gen Architecture}
        \label{sec:FE-Gen arch}
            Following recent practices~\cite{fetalflex,wu2025mrgen,guo2025maisi}, a straightforward approach is to adapt SD for fetal ultrasound synthesis. However, we notice two limitations during implementation: ({1}) {\it Histogram distribution drift}: generated images tend to be low-contrast and blurry, with histogram distributions deviating from real ultrasound as shown in Fig.~\ref{fig:hist-distribution}, which remains under-explored in previous literature; ({2}) {\it Sub-optimal image-mask alignment}: Recent studies~\cite{fetalflex} neglect quantitative measurement of spatial alignment between generated images and conditions, and conventional binary or RGB embeddings for spatial guidance may interfere with anatomy semantics and degrade spatial control precision.
            To address these issues, we propose \textbf{FE-Gen}, which is tailored for high-fidelity and spatially consistent fetal ultrasound synthesis with two key designs: 1) learnable semantic queries, and 2) histogram distribution regularization. Details are as follows. 
    
            {\it Learnable semantic queries.} 
            Considering plain color embeddings lack explicit semantic meaning and may cause ambiguity, particularly between categories with similar colors, we instead utilize learnable semantic queries inspired by SP-Ctrl~\cite{spctrl}.
            As shown in Fig.~\ref{fig:FE-Gen}(a), each anatomical category $k\in K$ is initialized with a fixed random vector $\boldsymbol{e}_k^0\in\mathbb{R}^{1\times C}$, where $C$ is the embedding dimension. An encoding module $\mathbb{E}(\cdot;\phi)$, which consists three stacked linear blocks parameterized by $\phi$, transforms it into a learnable semantic query $\boldsymbol{e}_k = \mathbb{E}(\boldsymbol{e}_k^0; \phi)$, $\boldsymbol{e}_k\in \mathbb{R}^{1\times C}$. 
            The condition mask ${\mathcal{M}}\in \mathbb{R}^{H\times W\times C}$ is then constructed by assigning the learned queries $\{{\boldsymbol{e}}_k\}_{k=1}^K$ to each pixel according to category-index mask $\mathcal{M}_{idx}=\{m^\prime_{i,j}=k\}_{H\times W\times 1}$, formulated as, ${\mathcal{M}} = \{{m}_{i,j}=\boldsymbol{e}_{k} | m^\prime_{i,j}=k, m^\prime_{i,j}\in \mathcal{M}_{idx}\}_{H\times W\times C}$.
            These semantic queries $\{\boldsymbol{e}_k\}_{k=1}^K$ are learned adaptively during the training of the diffusion model. 
    
            {\it Histogram distribution regularization.} To mitigate the histogram drift as shown in Fig.~\ref{fig:visualization} and Fig.~\ref{fig:hist-distribution} when adapting SD and ControlNet for ultrasound synthesis, we introduce an L1-based sorted loss for distribution regularization. Specifically, since diffusion models generate images through iterative denoising, we first use a DDIM scheduler~\cite{song2020denoising} to estimate the clean image latent $\hat{\boldsymbol{x}}_0$ from the current noisy latent ${\boldsymbol{x}}_t$ and predicted noise $\hat{\epsilon}_\theta$ at timestep $t$, which is formulated as, 
            \begin{equation}
                \hat{\boldsymbol{x}}_0 = \frac{\boldsymbol{x}_t-\sqrt{1-\overline{\alpha}_t}\hat{\epsilon}_\theta(\boldsymbol{x}_t,t)}{\sqrt{\overline{\alpha}_t}}, 
                \label{eq:x0_prediction}
            \end{equation}
            where $\hat{\epsilon}_\theta(\boldsymbol{x}_t,t)$ is the predicted noise at timestep $t$, and $\alpha_t$ is decided by the noise scheduler. 
            Then, the histogram regularization $\mathcal{L}_{hist}$ is calculated by a sorted L1 loss between the clean latent $\boldsymbol{x_0}$ and the estimated $\hat{\boldsymbol{x}}_0$, formulated as, 
            \begin{align}
                \mathcal{L}_{hist} = \frac{1}{N} \sum_{i=1} &|\hat{{x}}_0^{(i)} - {x}_0^{(i)} |, \ x^{(i)}\in \texttt{sort}(\boldsymbol{x})
                \label{eq:hist_reg}
            \end{align}
            where $\texttt{sort}(\boldsymbol{x}) = [x^{(1)}, \cdots,x^{(N)}]$, $x^{(i)}$ denotes sorted pixel values and $x^{(i)}\leq x^{(j)}$, if $i\leq j$. 
            Validations see Sec.~\ref{sec:ablation}. 

        \subsubsection{Multi-step Training} 
        \label{sec:three-step training}
        We employ a three-step fine-tuning strategy to adapt SD to fetal heart ultrasound synthesis, including: (1) {\it SD-VAE adaptation}, (2) {\it SD-UNet fine-tuning}, and (3) {\it FE-Gen training}, inspired by existing adaptation paradigms~\cite{guo2025maisi, wu2025mrgen}. Fig.~\ref{fig:FE-Gen} shows the whole training process.

        {\it Step-1: SD-VAE Adaptation.} For latent diffusion models, the reconstruction quality of VAE decides the upper bound for image fidelity. SD-VAE, pretrained on natural images, is underrepresented for fetal echocardiography and loses fine details, consistent with observations~\cite{chen2023textdiffuser} when adapting SD to specific domain. Therefore, we fine-tune VAE decoder on \texttt{WHU-4CH-8k} with the reconstruction loss $L_{rec}$ with MSE, perceptual loss $L_{p}$ with LPIPS, and GAN loss $L_{GAN}$, where
        \begin{equation}
            \mathcal{L}_{VAE}=L_{rec} + \lambda_1 L_{p} + \lambda_2 L_{GAN}, 
        \end{equation}
        where loss weights $\{\lambda_i\}_{i=1}^2$ set empirically following \cite{rombach2022stablediffusion}. We freeze the VAE encoder to keep the latent space unchanged. Table~\ref{tab:ablation_generation} verifies this adaptation with improved performance. 
        
        {\it Step-2: SD-UNet Fine-tuning.} Since SD, pretrained on natural images, cannot directly produce reasonable and realistic fetal echocardiography and exhibits large distribution gaps, we fine-tune SD-UNet with LoRA~\cite{hu2022lora} on \texttt{WHU-4CH-8k} for ultrasound synthesis, using a combination of denoising loss and histogram regularization, formulated as, 
        \begin{equation}
            \label{eq:unet}
            \mathcal{L}_{UNet} = \mathcal{L}_{denoise} + \omega_{h} \cdot \mathcal{L}_{hist}, 
        \end{equation}
        where $\omega_{h}$ denotes the regularization weight. We use a fixed text prompt $\mathbf{c}_t$ as ``{\it A realistic ultrasound image, fetal cardiac, standard four-chamber view.}''. Table~\ref{tab:ablation_generation} validates the effectiveness of $\mathcal{L}_{hist}$ in image fidelity. The fine-tuned SD can generate fetal ultrasound images that better match the real data distribution, serving as the basis for controllable generation. 

        {\it Step-3: FE-Gen Training.} We then train FE-Gen on image-mask pairs from \texttt{WHU-4CH-Seg517}, initialized with the fine-tuned SD-VAE and SD-UNet, to realize mask-guided fetal heart ultrasound generation, optimized with the same objective as Eq.~\ref{eq:unet}. The embedding module $\mathbb{E}(\cdot;\phi)$ and learnable semantic queries $\{\mathbf{e}_k\}_{k=1}^K$ are jointly optimized during the training. Empirically, jointly fine-tuning SD-UNet LoRA with FE-Gen yields better performance. Sec.~\ref{sec:ablation} further investigates the effects of training data scale, classifier-free guidance, and conditioning scale, providing practical guidelines for model configuration. Notably, FE-Gen introduces no additional inference cost compared with the baseline ControlNet.

\section{Experiments}
\label{sec:experiments}
    \subsection{Experimental Setup}
    \label{sec:settings}
        
        \renewcommand\arraystretch{1.0}
        \begin{table*}[t]
            \centering
            \caption{Comparison of fetal ultrasound synthesis methods. The best and second-best results are denoted by {\bf bold} and \underline{underline}. }
            \label{tab:generation performance}
            \scalebox{1.0}{
            \begin{tabular}{c|c|ccc|cccc|c}
                 \Xhline{1px}
                 \multirow{2}{*}{Model} & \multirow{2}{*}{Condition} & \multicolumn{8}{c}{Metrics} \\ 
                 \cline{3-10}
                 ~ & ~ & \multicolumn{1}{c}{KL$\downarrow$} & \multicolumn{1}{c}{Skewness} & \multicolumn{1}{c|}{STD} & LPIPS$\downarrow$ & KID$\downarrow$ & \multicolumn{1}{c}{FID$\downarrow$} & \multicolumn{1}{c|}{CLIPScore} & DSC$\uparrow$  \\ 
                 \Xhline{0.7px}
                 \rowcolor{gray!20} \textcolor{black}{Real Data} & \textcolor{black}{—} & \textcolor{black}{0.0000} & \textcolor{black}{0.64} & \textcolor{black}{50.89} & \textcolor{black}{0.0000} & \textcolor{black}{0.0000} & \textcolor{black}{0.00} & \textcolor{black}{35.25} & \textcolor{black}{80.75} \\ 
                 \hline
                 DreamBooth & text & 0.4181 & 0.07 & {\bf 49.81} & 0.5742 & 0.0822 & 170.71 & 38.82 & 11.58 \\ 
                 SD-LoRA & text & 0.1497 & 0.81 & 40.26 & 0.5727 & {\bf 0.0382} & {\bf 147.42} & 31.24 & 10.72 \\ 
                 \hline
                 T2I-Adapter & mask & \underline{0.1169} & 1.04 & 40.26 & 0.4962 & 0.0911 & 180.57 & {\bf 36.69} & 72.05 \\ 
                 ControlNet & mask & 0.3047 & \underline{0.49} & 35.76 & \underline{0.4570} & 0.0575 & 161.10 & 37.64 & \underline{75.68} \\ 
                 \cellcolor{lightblue}{Ours} & \cellcolor{lightblue}{mask} & \cellcolor{lightblue}{\bf 0.0350} & \cellcolor{lightblue}{\bf 0.54} & \cellcolor{lightblue}{\underline{46.32}} & \cellcolor{lightblue}{\bf 0.4347} & \cellcolor{lightblue}{\underline{0.0423}} & \cellcolor{lightblue}{\underline{148.13}} & \cellcolor{lightblue}{\underline{36.83}} & \cellcolor{lightblue}{\bf 79.52} \\
                \Xhline{1px}
            \end{tabular}}
            \vspace{-0.1cm}
        \end{table*}         
        
        \begin{figure*}[t]
            \centering
            \includegraphics[width=0.94\linewidth]{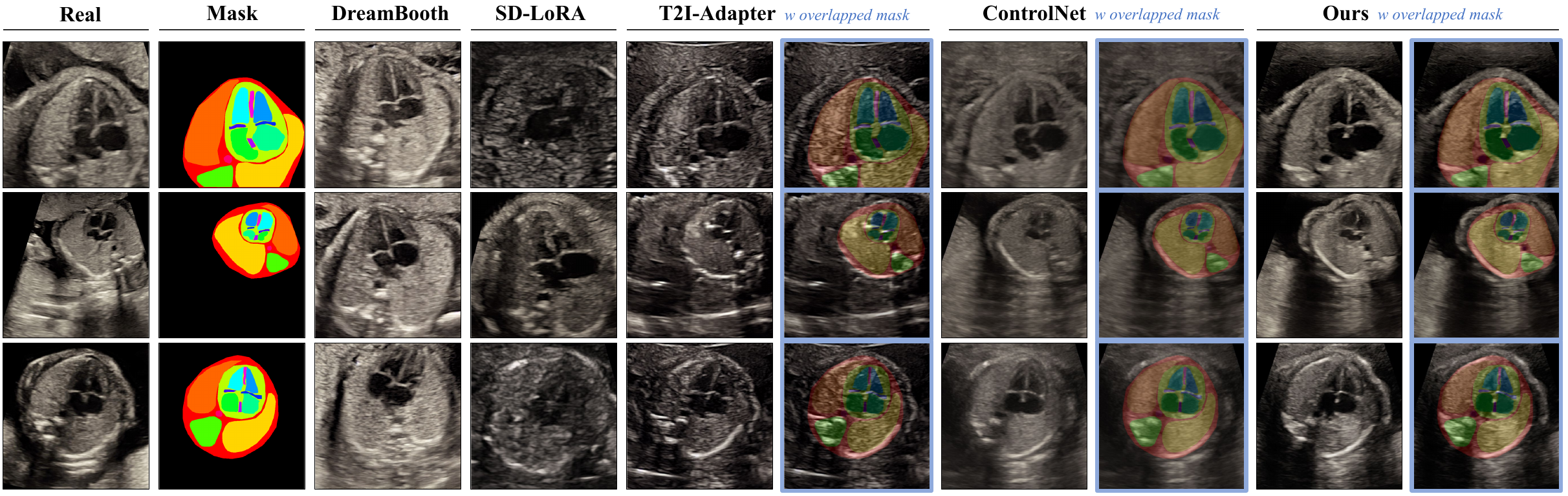}
            \caption{Comparisons of fetal heart ultrasound images generated by different methods with overlapped semantic masks for visualization.}
            \vspace{-0.8em}
            \label{fig:visualization}
        \end{figure*}
    
        \subsubsection{Datasets} 
        Due to the lack of publicly available datasets for fetal heart ultrasound, we collect two in-house datasets with clinician assistance as described in Sec.~\ref{sec:data_collection}, including: 1) \texttt{WHU-4CH-8k}, split into 6,466 training and 1,508 test images for fine-tuning SD-VAE and SD-UNet, and 2) \texttt{WHU-4CH-Seg517}, with 14 anatomical annotations, split into 417 training and 100 test images for training FE-Gen. 
        
        To evaluate synthesized data on real-world CHD analysis tasks, we select three heterogeneous CHD subtypes---EbA, TAPVC, and VSD---with different incidence rates and prognoses as representatives. We collect extra real positive cases for each subtype with the assistance of experts and consider three distinct settings for evaluation, including:

        ({1}) {\it Zero-shot classification}: 
        For rare CHDs with poor prognoses and insufficient training positives, we construct two real-world test sets for EbA~($0.53\%$ of CHDs) and TAPVC~($1.50\%$ of CHDs) referring to clinical reports. \texttt{CLS-EbA-test84} and \texttt{CLS-TAPVC-test80} consist of 42/40 normal and 42/40 positive cases from 5 and 7 patients separately. It represents scenarios where no real positives are available for training. 
        
        ({2}) {\it Lesion detection}: 
        For more prevalent CHD subtypes like VSD ($35.57\%$ of CHDs), where a few positives can be collected with effort, we construct \texttt{DET-VSD-908}, with a training set ($760$ negative/ $28$ positive) and a test set ($60$ negative/ $60$ positive) following \cite{li2024application}, to evaluate the effectiveness of synthesized samples under severe class imbalance scenarios.
        
        ({3}) {\it Anatomy segmentation}: 
        We evaluate synthesized normal samples for fetal cardiac anatomy segmentation using \texttt{WHU-4CH-Seg517}, where predicted masks can be used for biometric assessment and indicate possible pathologies. To avoid data leakage, data synthesis and augmentation use only automatically detected semantic masks from the \texttt{DET-VSD-908} training set.
        No test images or masks are used to train generative models to ensure fair evaluations. 

        \begin{figure}[t]
            \centering
            \includegraphics[width=0.82\linewidth]{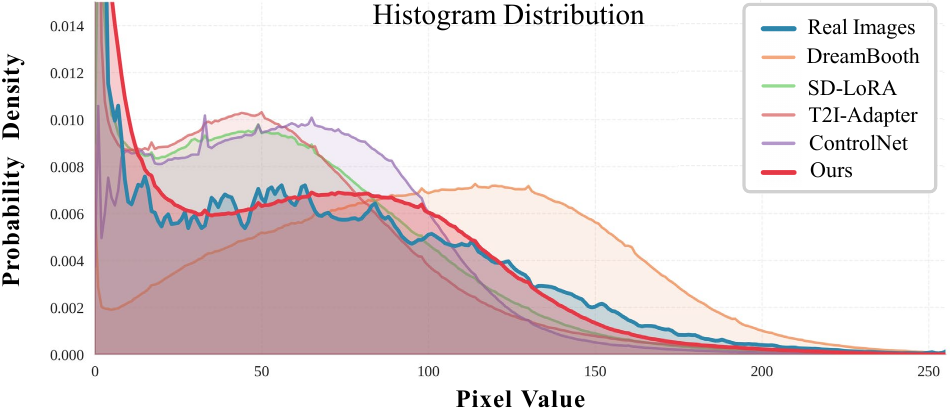}
            \caption{Histogram distribution comparison of synthesized images.}
            \vspace{-0.3em}
            \label{fig:hist-distribution}
        \end{figure}

        \begin{figure}[t]
            \centering
            \includegraphics[width=0.82\linewidth]{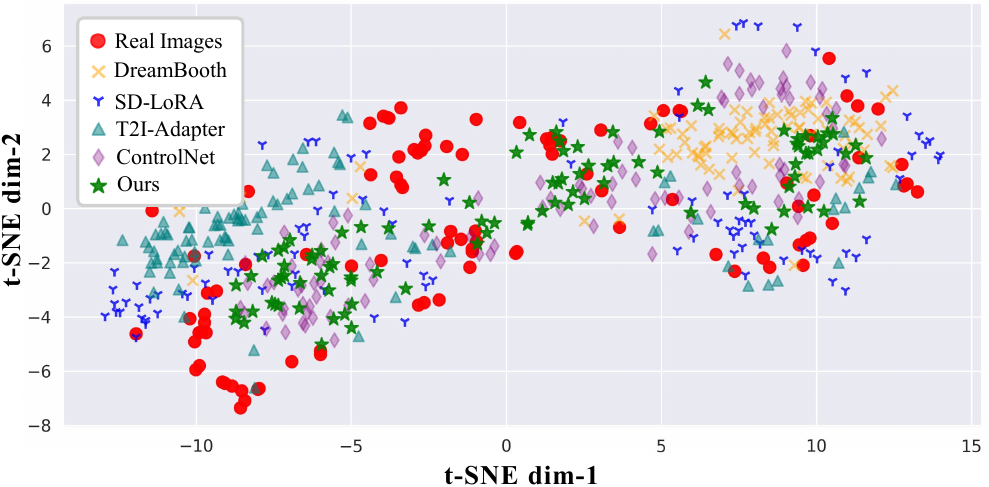}
            \caption{Data distribution visualization via t-SNE with FetalCLIP.}
            \vspace{-1.6em}
            \label{fig:tsne}
        \end{figure}

        \subsubsection{Metrics} 
        We evaluate fetal echocardiography generation from three aspects: (1) {Reconstruction}: {\it PSNR} and {\it SSIM} for VAE reconstruction quality; (2) {Image fidelity}: {\it FID} and {\it KID} measure feature distributions, {\it LPIPS} evaluate perceptual similarity, and {\it KL divergence}, {\it Skewness}, and {\it Standard Deviation} quantify histogram distribution differences. We further compute {\it CLIP-Score} using pretrained FetalCLIP~\cite{maani2026fetalclip} with fixed prompts to assess image-text distribution; (3) {Image-mask alignment}: Dice Similarity Coefficient~({\it DSC}) of synthesized images, estimated by an nnUNet model pre-trained on the \texttt{WHU-4CH-Seg517} training set. 
        For downstream tasks, we use standard metrics: {\it Accuracy, Recall, Precision, AUC} for classification; {\it mAP50} for detection; {\it Precision, Recall, DSC, mIoU} for segmentation. Each experiment is repeated three times or with $5$-fold cross-validation for robust evaluation.
        
        \subsubsection{Implementation}
        All experiments are conducted with PyTorch on four NVIDIA 3090 GPUs. FE-Gen is built on SD v1.5 and ControlNet.
        For SD-VAE adaptation, we fine-tune the decoder for 20 epochs with a batch size of 4 and a learning rate of $1e$$-$$5$.
        For SD-UNet adaptation, we adopt a LoRA rank of 8 for fine-tuning 2,000 epochs with a learning rate of $1e$$-$$4$ and a batch size of 16. The noise offset is set to 0.1~\cite{lin2024common} to alleviate contrast flaws, and we randomly drop $30\%$ prompts to support classifier-free guidance~(CFG). 
        For FE-Gen, ControlNet is initialized from the UNet encoder and trained for 2,000 epochs with a batch size of 16 and a learning rate of $1e$$-$$5$. The regularization weight $\omega_{h}$ is set to $0.1$ (see Sec.~\ref{sec:disc-regweight}). During inference, we adopt CFG$=$$2.0$ and conditioning scale $\lambda$$=$$1.0$ (see Sec.~\ref{sec:disc-gen-hyperparam}). 
        For downstream tasks, we use \texttt{ConvNeXt} for zero-shot classification, \texttt{Yolo-v5s} for lesion detection, and \texttt{nnUNet} for fetal heart anatomy segmentation following \cite{sarker2025harmonicechonet,li2024application} with default settings. Sec.~\ref{sec:exp_downstream} provides more details. All experiments are conducted with an image resolution of $512$$\times$$512$ unless otherwise specified. 

    \subsection{Experimental Results}
    \label{sec:main-exp}
        \subsubsection{Image Generation}
        \label{sec:exp_generation}
\begin{figure}[t]
            \centering
            \includegraphics[width=0.88\linewidth]{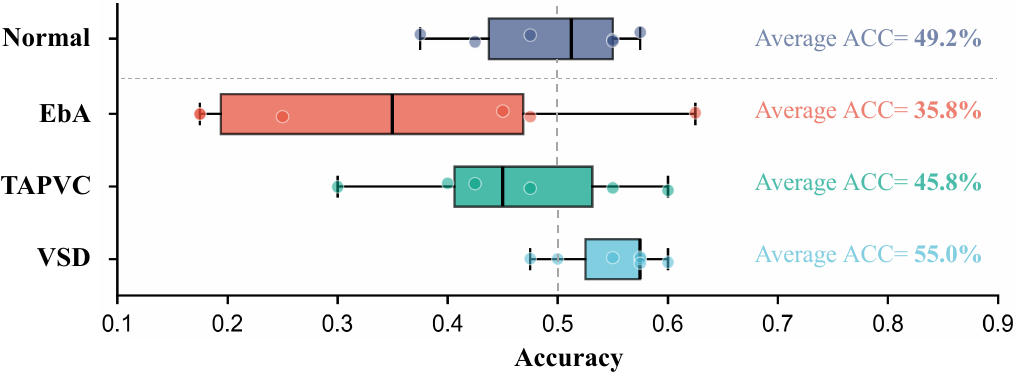}
            \vspace{-0.4em}
            \caption{Expert evaluation results on {\it real or fake} classification with real and our synthesized samples .}
            \vspace{-1.4em}
            \label{fig:user study}
        \end{figure}

        We compare FE-Gen with representative image synthesis methods in Table~\ref{tab:generation performance}: (1) SD-adaptation methods, {\it i.e.,} DreamBooth~\cite{ruiz2023dreambooth} and LoRA~\cite{hu2022lora}, and (2) spatially controllable generators, {\it i.e.}, T2I-Adapter~\cite{mou2024t2i} and ControlNet~\cite{zhang2023controlnet}. Our baseline is ControlNet~({\it CtrlNet}). 
        Notably, text-guided-only methods lack explicit spatial control and cannot follow masks. Our method achieves competitive FID and the lowest LPIPS, and surpasses CtrlNet by $3.96\%$ DSC, demonstrating superior image-mask alignment for subsequent counterfactual synthesis. Our method also achieves the lowest KL divergence, while others exhibit low standard deviations or large skewness gaps, suggesting reduced image contrast.
        Qualitative comparisons in Fig.~\ref{fig:visualization}, histogram distributions in Fig.~\ref{fig:hist-distribution}, and FetalCLIP~\cite{maani2026fetalclip} feature distributions in Fig.~\ref{fig:tsne} show that our synthesized data exhibit the closest distribution to real ones.        
        Furthermore, Fig.~\ref{fig:user study} reports the user study results with $6$ clinicians with an average 12-year experience in fetal ultrasound. For 20 real and 20 synthesized images of each type (Normal, VSD, EbA, and TAPVC), clinicians are asked to classify them as real or synthesized independently. The overall average accuracy is $46.4\%$, comparable to the $50\%$ chance level, indicating it is hard to distinguish synthesized samples from real data, reflecting our high perceptual fidelity.

        \subsubsection{Downstream Tasks} 
        \label{sec:exp_downstream}
        We evaluate synthesized samples on three CHD analysis tasks: zero-shot EbA/TAPVC classification, VSD detection, and fetal cardiac anatomy segmentation.

        {\it Zero-shot classification.} We synthesize EbA and TAPVC samples following Fig.~\ref{fig:FreePos} and train ConvNeXt for 200 epochs on $200$ positives ({\it Syn.}) and $200$ negatives ({\it Real or Syn.}), and evaluate on \texttt{CLS-EbA-test84} and \texttt{CLS-TAPVC-test80}. 
        As listed in Table~\ref{tab:zero-shot cls}, models trained entirely on synthesized data generalize well to real test images and even outperform those trained with real negatives, possibly because models overfit low-level features for distinguishing real and synthetic data rather than CHD-specific features. We also outperform CtrlNet, achieving $93.25\%$ ACC on EbA and $72.92\%$ ACC on TAPVC, owing to superior fidelity and spatial alignment. 
        
        {\it Lesion detection.} For VSD detection, we generate pseudo-positives from masks of the \texttt{WHU-4CH-Seg517} training set, apply mask deformation, and filter low-quality samples, yielding $\sim$$276$ synthesized positives for augmentation. We train Yolo-v5s for 250 epochs. As shown in Table~\ref{tab:vsd detection}, Pos4CHD improves the average mAP50 by $6.95\%$ over the real data, also $3.41\%$ higher than CtrlNet. FE-Gen retains $66.2\%$~($276/417$) generated samples after quality control, much higher than the $26.4\%$ of CtrlNet, exhibiting better data efficiency. 

        {\it Anatomy Segmentation.} 
        We evaluate Pos4CHD-synthesized normal cases on fetal cardiac segmentation. 
        We train nnUNet for 200 epochs on \texttt{WHU-4CH-Seg517} training set and explore its effects under various data scales in Table~\ref{tab:segmentation}. 
        We outperform CtrlNet under most settings. When only 20 real samples are available, adding synthesized data improves DSC by $10.49\%$. The performance gain diminishes as the real data scales over $200$ empirically, where further adding real data brings marginal increments. We filter out low-quality synthetic samples before training using a pretrained nnUNet with DSC$\leq$$80\%$ for rejection. Our method retains $57.5\%$~($437/760$) synthesized samples, higher than the $35.4\%$ utility of CtrlNet.

        \renewcommand\arraystretch{1.0}
        \begin{table}[t]
            \begin{center}
            \caption{
            Zero-shot classification for EbA and TAPVC. \texttt{Real} and \texttt{Syn} denote real and synthesized data. {\bf Bold} for the best result.
            }
            \label{tab:zero-shot cls}
            \scalebox{0.9}{
            \begin{tabular}{c|c|c|cccc}
            \Xhline{1.2pt}
            \multirow{2}{*}{Model} & \multicolumn{2}{c|}{Train Set} & \multirow{2}{*}{ACC(\%)} & \multirow{2}{*}{P(\%)} & \multirow{2}{*}{R(\%)} & \multirow{2}{*}{AUC(\%)} \\ 
            \cline{2-3}
            ~ & Neg. & Pos. & ~ & ~ & ~ & ~ \\ 
            \Xhline{1.2pt}
            \multicolumn{7}{c}{\textcolor{gray}{\textbf{EbA Classification}}} \\
            \hline
            \multirow{2}{*}{CtrlNet} & Real & Syn & 58.73\textcolor{gray}{\tiny $\pm$1.82} & 88.81\textcolor{gray}{\tiny $\pm$9.79} & 20.64\textcolor{gray}{\tiny $\pm$5.99} & 83.86\textcolor{gray}{\tiny $\pm$9.84} \\
            ~ & Syn & Syn & 91.27\textcolor{gray}{\tiny $\pm$1.82} & 94.11\textcolor{gray}{\tiny $\pm$1.09} & 88.09\textcolor{gray}{\tiny $\pm$4.76} & {\bf 96.09}\textcolor{gray}{\tiny $\pm$2.60} \\
            \hline
            \rowcolor{lightblue} ~ & Real & Syn & 71.43\textcolor{gray}{\tiny $\pm$1.19} & {\bf 100.00}\textcolor{gray}{\tiny $\pm$0.00} & 42.86\textcolor{gray}{\tiny $\pm$2.38} & 83.73\textcolor{gray}{\tiny $\pm$8.78} \\
            \rowcolor{lightblue} \multirow{-2}{*}{Ours} & Syn & Syn & {\bf 93.25}\textcolor{gray}{\tiny $\pm$0.69} & {93.61}\textcolor{gray}{\tiny $\pm$1.30} & {\bf 92.66}\textcolor{gray}{\tiny $\pm$0.17} & 95.03\textcolor{gray}{\tiny $\pm$0.92} \\
            \hline
            \multicolumn{7}{c}{\textcolor{gray}{\textbf{TAPVC Classification}}} \\
            \hline
            \multirow{2}{*}{CtrlNet} & Real & Syn & 53.33\textcolor{gray}{\tiny $\pm$2.60} & 95.24\textcolor{gray}{\tiny $\pm$8.25} & 7.5\textcolor{gray}{\tiny $\pm$6.61} & 75.12\textcolor{gray}{\tiny $\pm$9.32} \\
            ~ & Syn & Syn & 67.08\textcolor{gray}{\tiny $\pm$5.64} & 73.39\textcolor{gray}{\tiny $\pm$3.69} & 53.33\textcolor{gray}{\tiny $\pm$15.27} & 72.79\textcolor{gray}{\tiny $\pm$4.34} \\
            \hline
            \rowcolor{lightblue} ~ & Real & Syn & 62.92\textcolor{gray}{\tiny $\pm$1.44} & {\bf 100.00}\textcolor{gray}{\tiny $\pm$0.00} & 25.83\textcolor{gray}{\tiny $\pm$2.88} & 74.11\textcolor{gray}{\tiny $\pm$2.74} \\
            \rowcolor{lightblue}  \multirow{-2}{*}{Ours} & Syn & Syn & {\bf 72.92}\textcolor{gray}{\tiny $\pm$2.60} & 70.27\textcolor{gray}{\tiny $\pm$3.71} & {\bf 80.00}\textcolor{gray}{\tiny $\pm$2.50} & {\bf 77.54}\textcolor{gray}{\tiny $\pm$5.89} \\
            \Xhline{1.2pt}
            \end{tabular}}
            \end{center}

            \begin{center}
            \caption{Results of VSD detection. \texttt{Real} and \texttt{Syn} denote real and synthesized data. {\bf Bold} for the best result.}
            \label{tab:vsd detection}
            \scalebox{0.96}{
            \begin{tabular}{c|c|c|ccc}
            \Xhline{1.2pt}
            \multirow{2}{*}{Model} & \multicolumn{2}{c|}{Train Set} & \multicolumn{3}{c}{mAP50(\%)}  \\ 
            \cline{2-3}
            ~ & Neg & Pos & AVG & IVS & VSD \\ 
            \Xhline{1.2pt}
            — & Real & Real & 71.43\textcolor{gray}{\tiny $\pm$0.40} & 69.60\textcolor{gray}{\tiny $\pm$2.82} & 73.30\textcolor{gray}{\tiny $\pm$2.96} \\
            \hline
            CtrlNet & Real & Real+Syn & 74.97\textcolor{gray}{\tiny $\pm$0.65} & 73.55\textcolor{gray}{\tiny $\pm$2.92} & 76.32\textcolor{gray}{\tiny $\pm$2.21} \\
            \cellcolor{lightblue}{Ours} & \cellcolor{lightblue}{Real} & \cellcolor{lightblue}{Real+Syn} & \cellcolor{lightblue}{\bf 78.38\textcolor{gray}{\tiny $\pm$0.50}} & \cellcolor{lightblue}{\bf 75.68\textcolor{gray}{\tiny $\pm$2.25}} & \cellcolor{lightblue}{\bf 80.43\textcolor{gray}{\tiny $\pm$1.85}} \\
            \Xhline{1.2pt}
            \end{tabular}}
            \end{center}

            \begin{center}
            \caption{Fetal heart anatomy segmentation results of different data scales. {\bf Bold} for the best result.}
            \label{tab:segmentation}
            \scalebox{0.94}{
            \begin{tabular}{c|c|cccc}
            \Xhline{1.2pt}
            Scale & Aug & P(\%) & R(\%) & DSC(\%) & mIoU(\%) \\ 
            \Xhline{1.2pt}
            \multirow{3}{*}{0} & Baseline & — & — & — & — \\
            ~ & CtrlNet & 75.36\textcolor{gray}{\tiny $\pm$0.30} & 73.93\textcolor{gray}{\tiny $\pm$0.70} & 73.67\textcolor{gray}{\tiny $\pm$0.55} & 60.02\textcolor{gray}{\tiny $\pm$0.54} \\
            \rowcolor{lightblue} ~ & Ours & {\bf 76.91}\textcolor{gray}{\tiny $\pm$0.80} & {\bf 76.72}\textcolor{gray}{\tiny $\pm$0.58} & {\bf 75.94}\textcolor{gray}{\tiny $\pm$0.32} & {\bf 63.15}\textcolor{gray}{\tiny $\pm$0.48} \\
            \hline
            \multirow{3}{*}{20} & Baseline & 71.01\textcolor{gray}{\tiny $\pm$1.97} & 68.99\textcolor{gray}{\tiny $\pm$1.46} & 66.94\textcolor{gray}{\tiny $\pm$1.64} & 54.43\textcolor{gray}{\tiny $\pm$1.89} \\
            ~ & CtrlNet & 77.90\textcolor{gray}{\tiny $\pm$0.24} & 76.62\textcolor{gray}{\tiny $\pm$0.18} & 76.15\textcolor{gray}{\tiny $\pm$0.16} & 63.51\textcolor{gray}{\tiny $\pm$0.24} \\
            \rowcolor{lightblue} ~ & Ours & {\bf 79.15}\textcolor{gray}{\tiny $\pm$0.22} & {\bf 78.06}\textcolor{gray}{\tiny $\pm$0.45} & {\bf 77.43}\textcolor{gray}{\tiny $\pm$0.28} & {\bf 65.43}\textcolor{gray}{\tiny $\pm$0.38} \\
            \hline
            \multirow{3}{*}{50} & Baseline & 77.93\textcolor{gray}{\tiny $\pm$0.41} & 75.37\textcolor{gray}{\tiny $\pm$0.86} & 74.61\textcolor{gray}{\tiny $\pm$0.68} & 62.91\textcolor{gray}{\tiny $\pm$0.80} \\
            ~ & CtrlNet & 79.58\textcolor{gray}{\tiny $\pm$0.29} & 78.57\textcolor{gray}{\tiny $\pm$0.35} & 78.12\textcolor{gray}{\tiny $\pm$0.25} & 66.19\textcolor{gray}{\tiny $\pm$0.29} \\
            \rowcolor{lightblue} ~ & Ours & {\bf 79.93}\textcolor{gray}{\tiny $\pm$0.33} & {\bf 79.32}\textcolor{gray}{\tiny $\pm$0.26} & {\bf 78.48}\textcolor{gray}{\tiny $\pm$0.27} & {\bf 66.88}\textcolor{gray}{\tiny $\pm$0.34} \\
            \hline
            \multirow{3}{*}{100} & Baseline & 80.84\textcolor{gray}{\tiny $\pm$0.18} & 78.89\textcolor{gray}{\tiny $\pm$0.23} & 78.59\textcolor{gray}{\tiny $\pm$0.14} & 67.38\textcolor{gray}{\tiny $\pm$0.27} \\
            ~ & CtrlNet & 80.57\textcolor{gray}{\tiny $\pm$0.23} & {\bf 80.07}\textcolor{gray}{\tiny $\pm$0.26} & 79.38\textcolor{gray}{\tiny $\pm$0.08} & 67.98\textcolor{gray}{\tiny $\pm$0.11} \\
            \rowcolor{lightblue} ~ & Ours & {\bf 81.02}\textcolor{gray}{\tiny $\pm$0.16} & {79.99}\textcolor{gray}{\tiny $\pm$0.32} & {\bf 79.43}\textcolor{gray}{\tiny $\pm$0.27} & {\bf 68.21}\textcolor{gray}{\tiny $\pm$0.31} \\
            \hline
            \multirow{3}{*}{200} & Baseline & {\bf 81.69}\textcolor{gray}{\tiny $\pm$0.25} & 80.10\textcolor{gray}{\tiny $\pm$0.37} & 79.82\textcolor{gray}{\tiny $\pm$0.17} & 68.76\textcolor{gray}{\tiny $\pm$0.27} \\
            ~ & CtrlNet & 81.43\textcolor{gray}{\tiny $\pm$0.14} & {\bf 80.85}\textcolor{gray}{\tiny $\pm$0.19} & {\bf 80.14}\textcolor{gray}{\tiny $\pm$0.14} & {\bf 69.12}\textcolor{gray}{\tiny $\pm$0.21} \\
            \rowcolor{lightblue} ~ & Ours & 81.48\textcolor{gray}{\tiny $\pm$0.23} & 80.83\textcolor{gray}{\tiny $\pm$0.23} & 80.13\textcolor{gray}{\tiny $\pm$0.12} & {\bf 69.12}\textcolor{gray}{\tiny $\pm$0.24} \\
            \hline
            \multirow{3}{*}{400} & Baseline & {\bf 82.54}\textcolor{gray}{\tiny $\pm$0.19} & 81.19\textcolor{gray}{\tiny $\pm$0.19} & {\bf 80.89}\textcolor{gray}{\tiny $\pm$0.17} & 70.14\textcolor{gray}{\tiny $\pm$0.24} \\
            ~ & CtrlNet & 82.28\textcolor{gray}{\tiny $\pm$0.32} & 81.43\textcolor{gray}{\tiny $\pm$0.38} & 80.85\textcolor{gray}{\tiny $\pm$0.12} & {\bf 70.17}\textcolor{gray}{\tiny $\pm$0.11} \\
            \rowcolor{lightblue} ~ & Ours & 82.01\textcolor{gray}{\tiny $\pm$0.17} & {\bf 81.52}\textcolor{gray}{\tiny $\pm$0.25} & 80.72\textcolor{gray}{\tiny $\pm$0.13} & 70.04\textcolor{gray}{\tiny $\pm$0.13} \\
            \Xhline{1.2pt}
            \end{tabular}}
            \end{center}
            \vspace{-0.7cm}
        \end{table}

        \begin{figure}[t]
            \centering
            \includegraphics[width=0.99\linewidth]{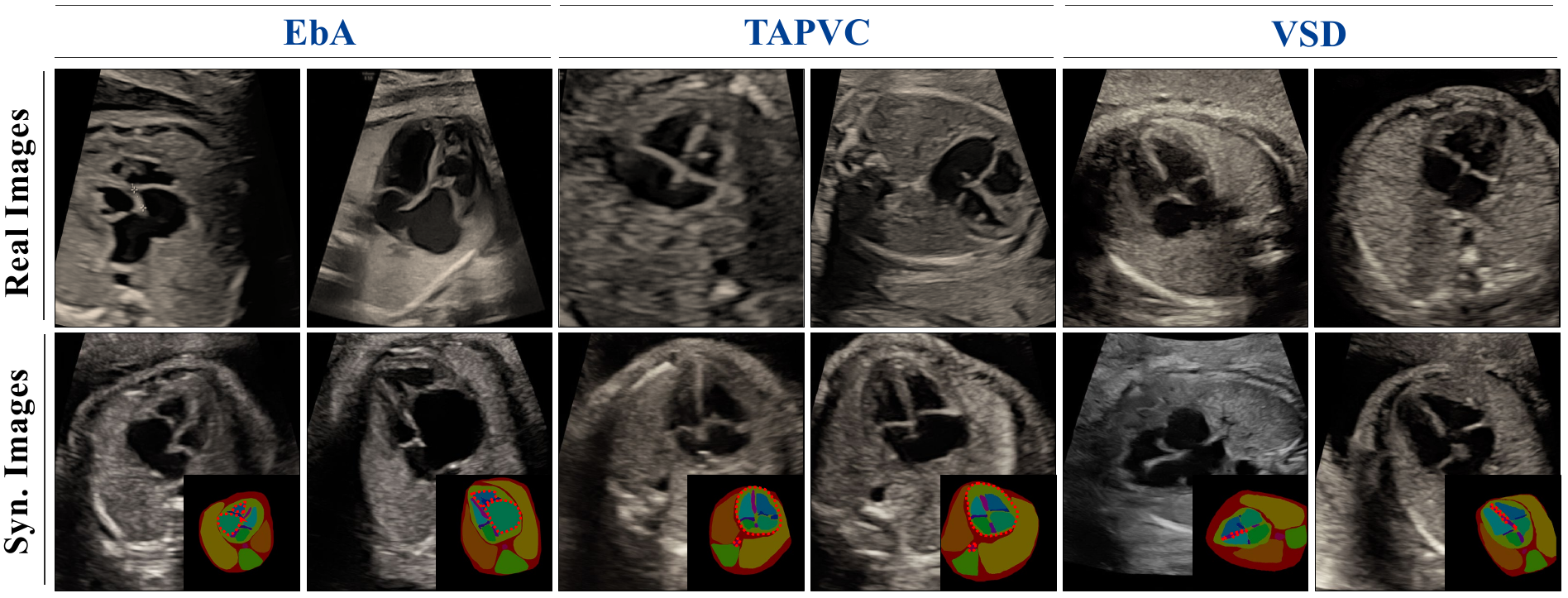}
            \caption{Comparison of synthesized CHD samples with real cases.}
            \vspace{-1.0em}
            \label{fig:synthesized positive samples}
        \end{figure}

    \subsection{Ablation Studies}
    \label{sec:ablation}
        
        \renewcommand\arraystretch{1.05}
        \begin{table}[t]
            \begin{center}
            \caption{Ablation studies for ultrasound generation.}
            \label{tab:ablation_generation}
            \scalebox{0.84}{
            \begin{tabular}{l|cccccc}
            \Xhline{1.2pt}
            Settings & \multicolumn{6}{c}{Metrics} \\ 
            \hline
            \multicolumn{7}{l}{\textcolor{gray}{\it Step-1: SD-VAE Adaptation}} \\
            \hline
            ~ & SSIM$\uparrow$ & PSNR$\uparrow$ & LPIPS$\downarrow$ & KL$\downarrow$ & KID$\downarrow$ & FID$\downarrow$ \\
            \hline
            \rowcolor{gray!20} Real Data & 1.0000 & inf & 0.0000 & 0.0000 & 0.0000 & 0.00 \\
            SD-VAE & 0.9306 & 35.18 & 0.0325 & 0.0180 & 0.0048 & 10.77 \\
            \quad {\it w fine-tune} & 0.9496 & 37.72 & 0.0168 & 0.0010 & 0.0008 & 4.83 \\
            \hline
            \multicolumn{7}{l}{\textcolor{gray}{\it Step-2: SD-UNet Fine-tuning}} \\
            \hline
            ~ & KL$\downarrow$ & Skew & STD & LPIPS$\downarrow$ & KID$\downarrow$ & FID$\downarrow$ \\
            \hline
            \rowcolor{gray!20} Real Data & 0.0000 & 0.76 & 51.52 & 0.0000 & 0.0000 & 0.00 \\
            SD-UNet~(LoRA) & 0.1159 &0.77 & 42.97 & 0.5872 & 0.0311 & 49.54 \\
            \quad {\it w $\mathcal{L}_{hist}$} & 0.0194 & 0.73 & 55.18 & 0.6055 & 0.0153 & 37.48 \\
            \hline
            \multicolumn{7}{l}{\textcolor{gray}{\it Step-3: FE-Gen Training}} \\
            \hline
            ~ & KL$\downarrow$ & Skew & STD & KID$\downarrow$ & FID$\downarrow$ & DSC$\uparrow$ \\
            \hline
            \rowcolor{gray!20} Real Data & 0.0000 & 0.64 & 50.89 & 0.0000 & 0.00 & 80.75 \\
            ControlNet & 0.3047 & 0.55 & 35.76 & 0.0575 & 161.10 & 75.68 \\
            \rowcolor{lightblue} Ours & 0.0350 & 0.54 & 46.32 & 0.0423 & 148.13 & 79.52 \\
            \quad {\it w/o} $\mathcal{L}_{hist}$ & 0.2822 & 0.48 & 35.10 & 0.0846 & 177.65 & 79.23 \\
            \quad {\it w/o step-2 finetune} & 0.0216 & 0.55 & 48.55 & 0.0618 & 162.96 & 79.10 \\
            \quad {\it w/o learned query} & 0.0312 & 0.68 & 46.30 & 0.0428 & 148.97 & 74.55 \\
            \Xhline{1.2pt}
            \end{tabular}}
            \end{center}
            \vspace{-0.6cm}
        \end{table}
    
        \subsubsection{Effects of regularization $\mathcal{L}_{hist}$} As shown in Table~\ref{tab:ablation_generation} and Fig.~\ref{fig:hist-distribution}, the proposed histogram regularization effectively reduces the distribution gap between synthesized and real images during SD-UNet fine-tuning and FE-Gen training. It leads to lower FID and improves image fidelity, while the baseline shows much lower standard deviation than real data.

        \subsubsection{Effects of learnable semantic query} 
        Table~\ref{tab:ablation_generation} also validates the effectiveness of learnable semantic queries, improving DSC by $4.97\%$, which is critical for downstream applications. Fig.~\ref{fig:visualization} also shows better image-mask alignments.

        \subsubsection{Effects of training strategy} 
        Table~\ref{tab:ablation_generation} ablates the performance of multi-step fine-tuning described in Sec.\ref{sec:three-step training}. We validate that fine-tuning SD-VAE and SD-UNet both benefit the overall image fidelity of our FE-Gen. 

        \subsubsection{Effects of human filtering \& mask augmentation} 
        \label{sec:valid_filtering}
\begin{figure}[t]
            \centering
            \includegraphics[width=0.98\linewidth]{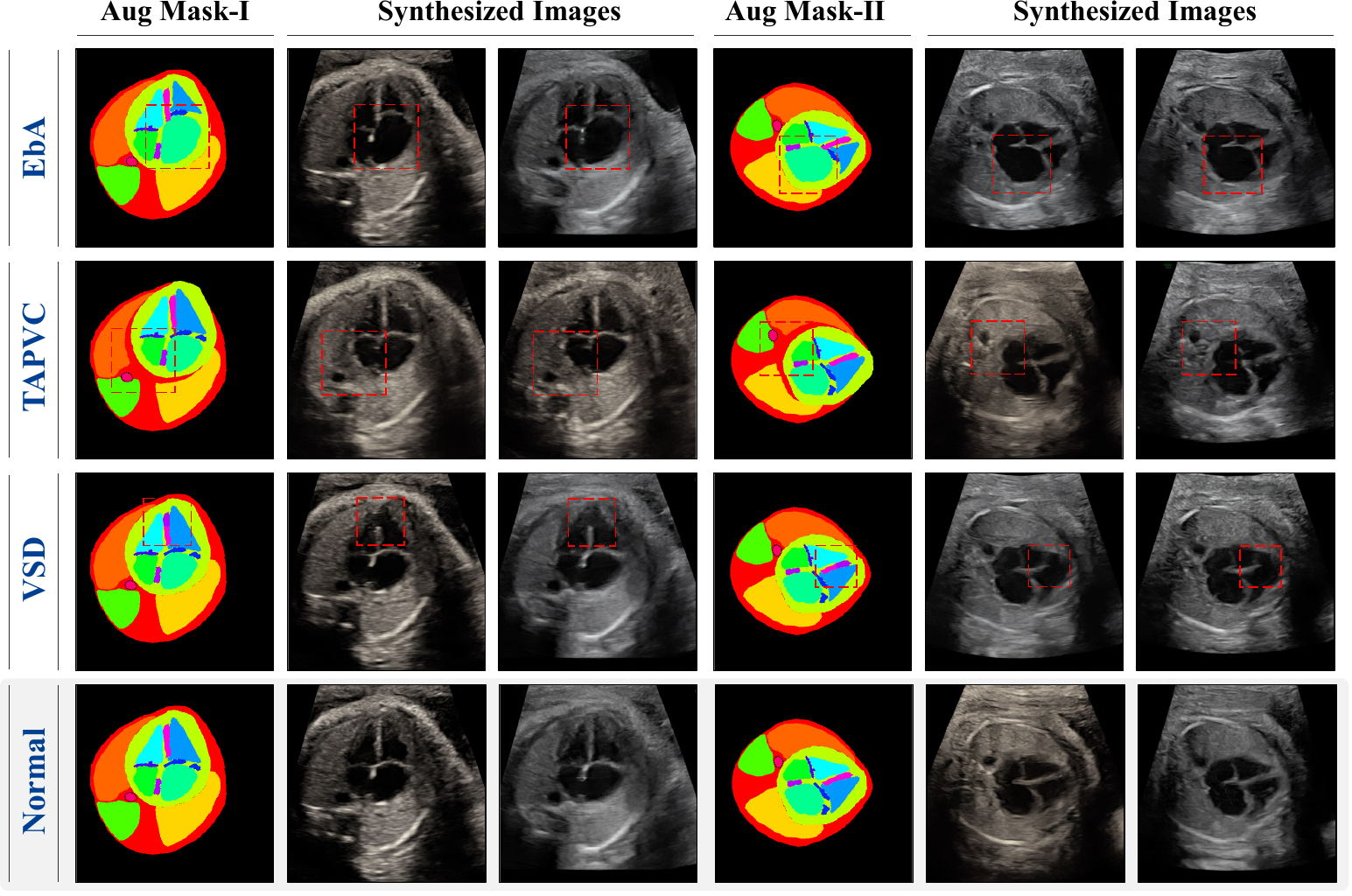}
            \caption{Diverse images synthesized from the same source mask.}
            \label{fig:diversity}
        \vspace{0.8em}

            \centering
            \includegraphics[width=0.86\linewidth]{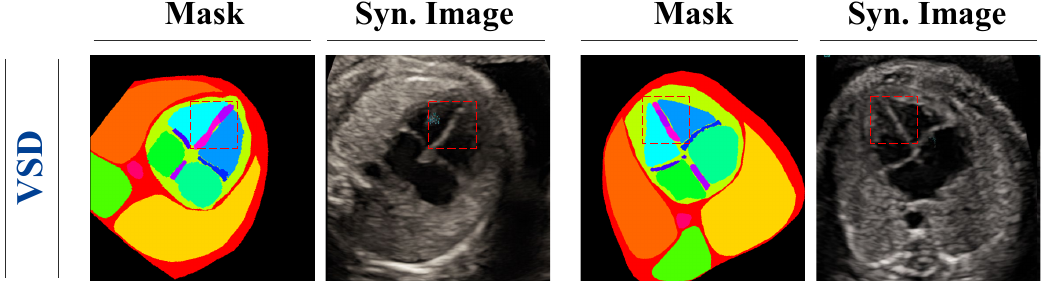}
            \caption{Examples of failure cases of Pos4CHD.}
            \vspace{-1.2em}
            \label{fig:failure case}
        \end{figure}

        Human filtering removes low-quality images that fail to exhibit reasonable CHD characteristics (see Fig.~\ref{fig:failure case}), thereby reducing noisy annotations for training. Mask augmentation improves mask diversity, leading to better sample diversity (see Fig.~\ref{fig:diversity}). Both strategies improve mAP50 in Table~\ref{tab:ablation_filter_aug}, and their combination increases $6.95\%$ on average and $7.13\%$ for VSD.
        
        \renewcommand\arraystretch{1.0}
        \begin{table}[t]
            \begin{center}
            \caption{Validation of data filtering and mask augmentation.}
            \label{tab:ablation_filter_aug}
            \begin{tabular}{c|cc|ccc}
            \Xhline{1.2pt}
            \multirow{2}{*}{Dataset} & Human & Mask & \multicolumn{3}{c}{mAP50} \\ 
            ~ & Filter & Aug & AVG & VSD & IVS \\ 
            \Xhline{1.2pt}
            Real & --- & --- & 71.43 & 73.30 & 69.60 \\
            \hline
            \multirow{4}{*}{+Syn} & $\times$ & $\times$ & 73.50 & 70.37 & 76.67 \\
            ~ & $\checkmark$ & $\times$ & 77.63 & 76.03 & \bf 79.27  \\
            ~ & $\times$ & $\checkmark$ & 77.17 & 79.93 & 74.43  \\
            ~ & $\checkmark$ & $\checkmark$ & \bf 78.38 & \bf 80.43 & 75.50  \\
            \Xhline{1.2pt}
            \end{tabular}
            \end{center}
            \vspace{-0.8em}
        \end{table}

        \begin{figure}[t]
            \centering
            \includegraphics[width=0.96\linewidth]{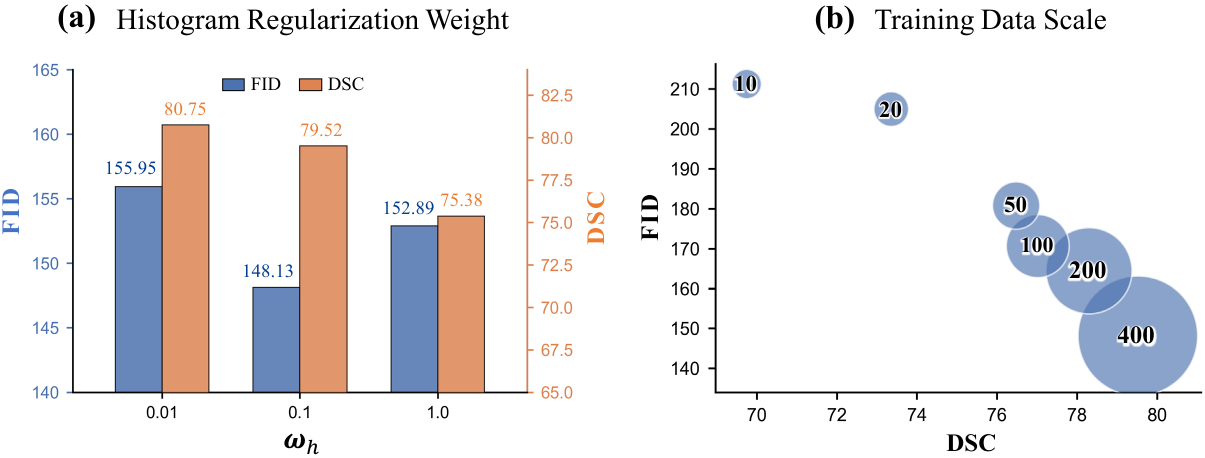}
            \vspace{-0.2em}
            \caption{Discussions on (a) regularization weight, (b) training data scale.}
            \vspace{0.8em}
            \label{fig:hist-reg and scale}

            \centering
            \includegraphics[width=0.96\linewidth]{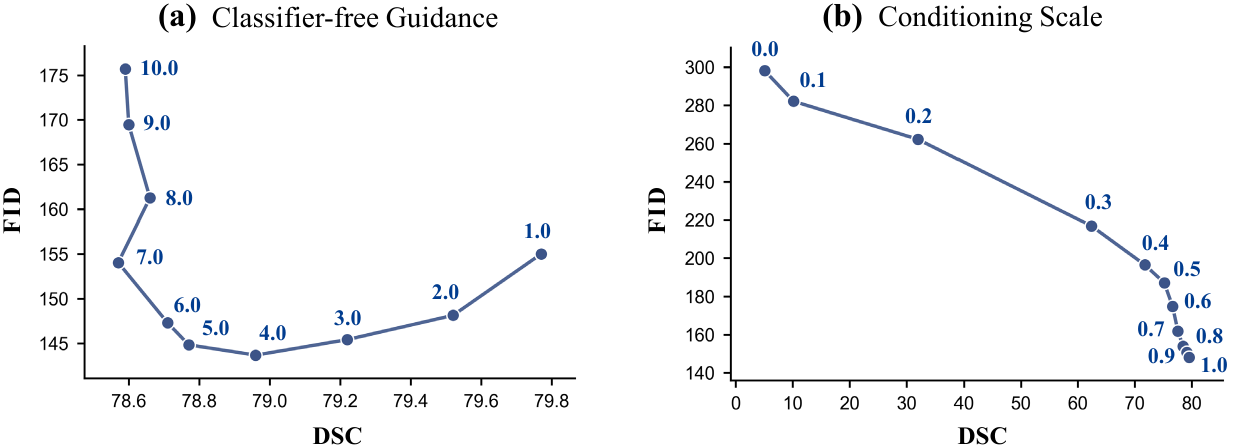}
            \vspace{-0.2em}
            \caption{Discussions on (a) CFG scale and (b) conditioning scale.}
            \vspace{0.8em}
            \label{fig:cfg_cond scale}
            
            \centering
            \includegraphics[width=0.96\linewidth]{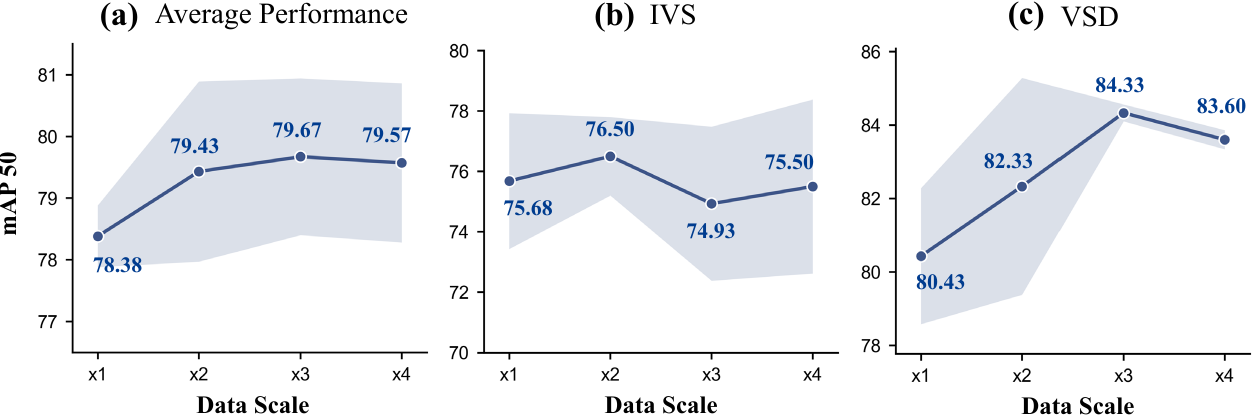}
            \vspace{-0.2em}
            \caption{Discussions on synthesized data scales for VSD detection.}
            \vspace{-0.8em}
            \label{fig:vsd detection data scale}
        \end{figure}
        
    \subsection{Discussions}
    \label{sec:discussion}

        \subsubsection{Discussions on regularization weight $\omega_h$}
        \label{sec:disc-regweight}
        Considering the trade-off between image fidelity~(FID) and spatial alignment~(DSC) in Fig.~\ref{fig:hist-reg and scale}(a), we set the weight $\omega_h$$=$$0.1$ in Eq.\ref{eq:hist_reg}. 

        \subsubsection{Discussions on training data scales for FE-Gen} Fig.~\ref{fig:hist-reg and scale}(b) indicates that, 
        increasing the data scale consistently reduces FID and improves DSC, with diminishing marginal gains, revealing the importance of training data scale.

        \subsubsection{Discussions on key synthesis hyperparameters}
        \label{sec:disc-gen-hyperparam}
        As shown in Fig.~\ref{fig:cfg_cond scale}, the CFG scale exhibits a trade-off between FID and DSC, and we set it to $2.0$ for better spatial alignment. Increasing conditioning scale $\lambda$ consistently lowers FID and improves DSC, with the best performance achieved at $\lambda=1.0$.

        \subsubsection{Discussions on augmentation data scale} 
        Using all $417$ masks from the \texttt{WHU-4CH-Seg517} training set, we synthesize pseudo-VSD samples by Pos4CHD as $\times1$ scale, and repeat this process to construct larger scales. As shown in Fig.~\ref{fig:vsd detection data scale}, the performance improves with more synthesized samples and gradually saturates at approximately $\times3$ scale. 

        \subsubsection{Discussions on synthesized sample diversity} 
        As shown in Fig.~\ref{fig:diversity}, Pos4CHD can synthesize diverse samples for CHD subtypes and normal cases, which arises from: (1) structural variations introduced by stochastic mask augmentation and counterfactual manipulation, and (2) appearance variations induced by the randomness of image generation.

        \subsubsection{Analysis of failure cases} 
        Fig.~\ref{fig:failure case} shows failure cases that do not faithfully exhibit CHD-specific characteristics. Most occur in pseudo-VSD samples, where the IVS remains visually continuous despite the defect specified by the counterfactual mask, attributed to: (1) the inherent stochasticity of diffusion sampling and (2) the difficulty of precise spatial control over subtle, localized structures. Although our method improves spatial alignment over the baseline, precise synthesis of subtle structures remains an important direction for future work.

\section{Conclusion}
\label{sec:conclusion}
\color{black}
    This paper presents Pos4CHD, a positive-free mask-guided counterfactual generation framework for fetal echocardiography, to alleviate CHD data scarcity. We first introduce FE-Gen for ultrasound generation, which mitigates histogram shift via regularization and improves spatial image-mask alignments via learnable semantic queries. Based on FE-Gen, Pos4CHD transforms normal masks into counterfactual ones and generates corresponding images, yielding pseudo-CHD image-mask pairs for downstream analysis. Extensive experiments on synthesized samples exhibit consistent improvements on real-world tasks including zero-shot classification, lesion detection, and anatomy segmentation, demonstrating the potential of synthesized data for automated fetal CHD diagnosis and analysis.

\section*{Acknowledgment}
    The numerical calculations are done on the supercomputing system of the Supercomputing Center of WHU.
    We used GPT solely for grammar checking and manuscript polishing throughout preparation, and all edits were revised by authors.

\bibliographystyle{IEEEtran}
\bibliography{reference}

\end{document}